\pdfoutput=1

\documentclass{bmvc2k}

\title{RED-Sphere: Hyperspherical Residual Edge Debiasing for Cross-Population Fundus Disease Domain Generalization}

\addauthor{Yan Lin}{y.lin64@newcastle.ac.uk}{1}
\addauthor{Ziheng Wang}{ziheng.wang@kaust.edu.sa}{2}
\addauthor{Shuang Chen}{shuang.chen@durham.ac.uk}{3}
\addauthor{Amir Atapour-Abarghouei}{amir.atapour-abarghouei@durham.ac.uk}{3}
\addauthor{Stephen McGough}{stephen.mcgough@newcastle.ac.uk}{1}

\addinstitution{
 Newcastle University\\
 Newcastle upon Tyne, UK
}
\addinstitution{
 King Abdullah University of\\
 Science and Technology (KAUST)\\
 Thuwal, Saudi Arabia
}
\addinstitution{
 Durham University\\
 Durham, UK
}

\runninghead{Lin et al.}{RED-Sphere}

\usepackage{xcolor}
\usepackage{booktabs}
\usepackage{multirow}
\usepackage{graphicx}
\usepackage{threeparttable}
\usepackage{amssymb}
\usepackage{algorithm}
\usepackage{algpseudocode}
\usepackage{makecell}
\usepackage{adjustbox}
\usepackage{subcaption}

\usepackage[table]{xcolor}
\usepackage{colortbl, xcolor, multirow, booktabs, adjustbox}
\definecolor{yellow}{RGB}{255, 255, 150}      %
\definecolor{lightblue}{RGB}{173, 216, 230}   %
\definecolor{lightred}{RGB}{255, 182, 193}    %
\definecolor{lightgreen}{RGB}{144, 238, 144}  %

\definecolor{greenbox}{RGB}{144, 238, 144}    %
\definecolor{redbox}{RGB}{255, 182, 193}      %
\definecolor{bluebox}{RGB}{135, 206, 235}     %
\definecolor{yellowbox}{RGB}{255, 255, 0}     %
\definecolor{posgreen}{RGB}{198, 239, 206}  %
\definecolor{negred}{RGB}{255, 199, 206}    %

\definecolor{crbox}{RGB}{245,238,255}

\begin{document}

\maketitle

\begin{abstract}

Medical image classifiers are often trained within one source population, yet clinical deployment requires robustness to patients whose appearance, acquisition style, and disease prevalence differ from the source cohort. Existing fairness and robustness methods often require group supervision or treat appearance variation as an undifferentiated nuisance, which is insufficient when population-correlated low-level cues and lesion evidence share edge and texture structure. 
We study a strict source only cross population setting, where external populations are unseen during optimization, validation, scheduling, hyperparameter selection, and model selection. 
We propose RED-Sphere, a plug-and-play robustness framework for image classification under unseen population shifts. 
This estimates shortcut-sensitive nuisance responses with an edge and feature energy prior, attenuates dominant responses through residual soft gating, regularizes masked nuisance views with counterfactual inspired consistency and separation losses, and predicts labels with normalized spherical prototypes. The framework promotes angular semantic evidence over source correlated activation magnitude while preserving lesion related structure. 
Although demonstrated on 2D Scanning Laser Ophthalmoscopy (SLO) fundus classification for Age-Related Macular Degeneration (AMD) and Diabetic Retinopathy (DR), RED-Sphere is not tied to retinal anatomy because the same principle can be adapted with modality specific nuisance priors to classification settings where appearance shortcuts and semantic evidence are entangled. 
Under a strict White-only Harvard-FairVision protocol, RED-Sphere improves held out macro-F1 across all 20 task and backbone comparisons, with average gains of 1.28 and 2.98 F1 points on AMD and DR. Improvements in AUC and PR-AUC, together with visual diagnostics, ablations, and sensitivity analyses, further support stronger external semantic alignment and more stable angular disease geometry.

\end{abstract}

\vspace{-0.5cm}
\section{Introduction}
\label{sec:intro}

\begin{figure}[t]
  \centering
  \includegraphics[width=0.7\linewidth]{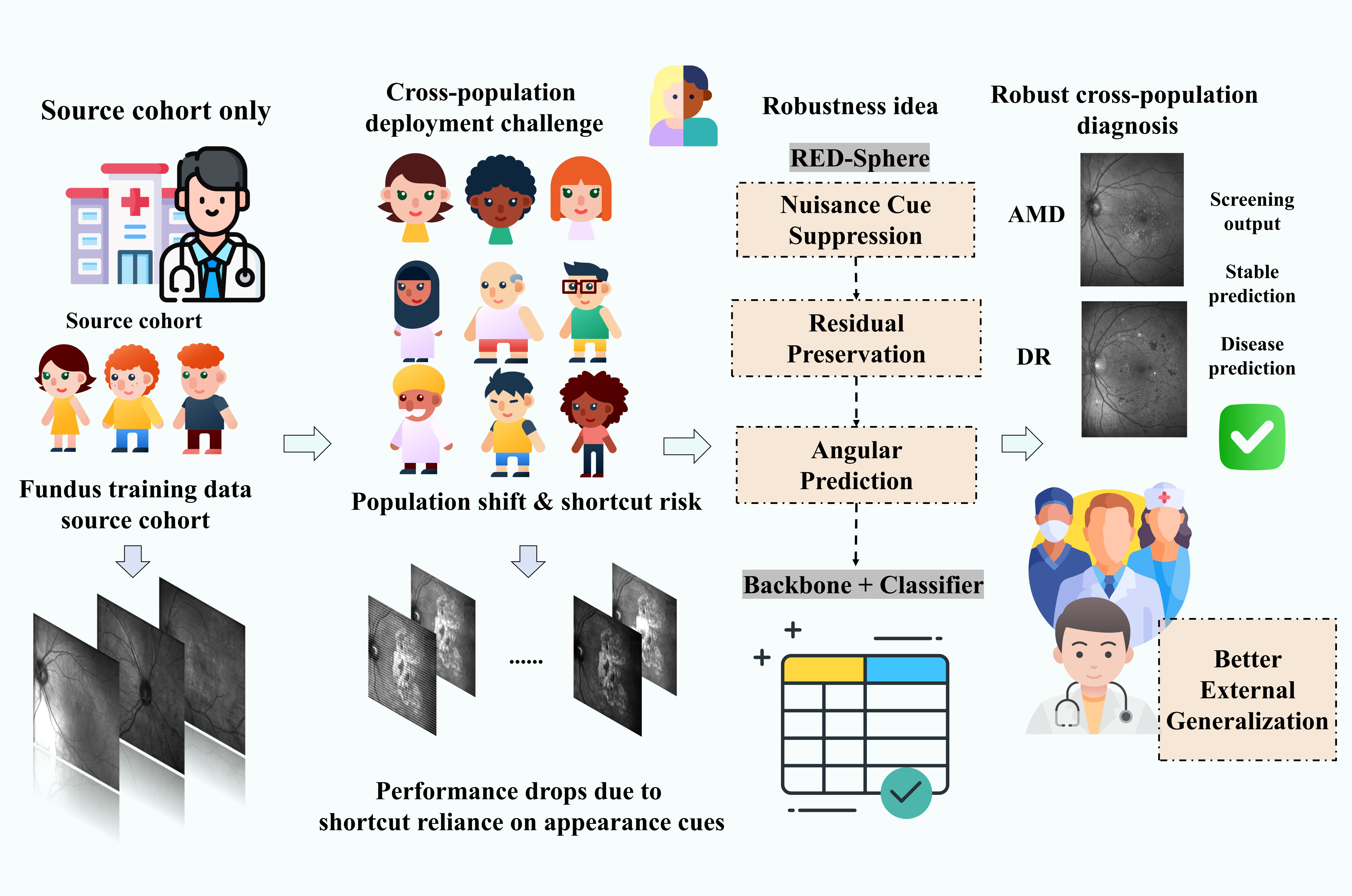}
  \vspace{0.2cm}
  \caption{Source only cross population fundus disease classification. Training and model selection use one source cohort while external populations remain unseen until testing.
  }
  \label{fig:first}
  \vspace{0.15cm}
\end{figure}

\vspace{-0.2cm}
Medical image classifiers are increasingly expected to operate beyond the population from which their training data are collected. A model trained within one source cohort may encounter patients with different acquisition protocols, appearance statistics, disease prevalence, and clinical workflows. Such source to external population shift is especially consequential in screening, where strong aggregate performance can coexist with poor sensitivity to clinically important minority categories. In ophthalmic imaging, Harvard-FairVision exposes demographic performance gaps in eye disease screening~\cite{luo2024fairvision}. Retinal images can also encode population information through pigmentation, luminance, color statistics, and vessel maps~\cite{coyner2023racialbias,rajesh2025retinalpigment}. A robust classifier should therefore avoid treating cohort correlated appearance as disease evidence.

Existing fairness and generalization methods address related forms of distribution shift, but source-only deployment remains a more restrictive and practically important setting, as external populations are unavailable during training, validation, and model selection. Demographic fairness objectives such as equality of opportunity~\cite{hardt2016equality}, counterfactual fairness~\cite{kusner2017counterfactual}, group distributional robustness~\cite{sagawa2020distributionally}, and adversarial debiasing~\cite{zhang2018mitigating} rely on sensitive labels, multiple observed groups, or group aware validation signals. 
Domain generalization benchmarks motivate evaluation on unseen distributions~\cite{gulrajani2021insearch}, but broad appearance perturbation does not decide which low level evidence should be suppressed and which low level evidence should be preserved.

We explore source-only medical generalization through nuisance aware robustness rather than access to target populations. The central hypothesis is that robust classification should estimate shortcut sensitive cues, preserve task relevant evidence through residual paths, regularize nuisance dominant views during training, and predict in normalized semantic geometry. The aim is not to erase appearance variation but to attenuate dominant nuisance responses while retaining medical structure that may share the same low level cues.

Fundus disease classification makes the tension concrete. 
In 2D SLO fundus images, cross population variation can appear through pigmentation, vessel contrast, illumination, texture, edge strength, and acquisition style~\cite{luo2024fairvision,coyner2023racialbias,jang2025fundusdg}. 
The same structures support diagnosis because drusen, hemorrhages, exudates, vessel abnormalities, and lesion boundaries contain local texture and edge patterns. Blind appearance removal can suppress clinically relevant evidence, so cross population robustness must weaken source correlated appearance reliance while preserving disease semantic evidence.

We propose RED-Sphere, a plug and play robustness module for source only medical image classification. 
RED-Sphere attaches to a visual backbone and transforms the final feature map into a residual gated semantic stream and a masked nuisance view. An edge and feature energy prior estimate shortcuts sensitive responses. Residual soft gating attenuates high mask responses while preserving original feature evidence. Counterfactual inspired nuisance regularization exposes the classifier to nuisance dominant views during training without generating images or using target population samples. A spherical prototype classifier predicts by angular similarity rather than raw feature magnitude, reducing reliance on illumination, contrast, image quality, or population correlated activation strength. This design makes RED-Sphere a general robustness principle rather than a modality-specific preprocessing strategy. By replacing the nuisance prior with task- and modality-dependent cues, the framework can be adapted to different medical imaging scenarios where shortcut-sensitive appearance and disease evidence are entangled.

\begin{figure}[t]
  \centering
  \begin{minipage}{0.35\linewidth}\centering
    \includegraphics[width=\linewidth]{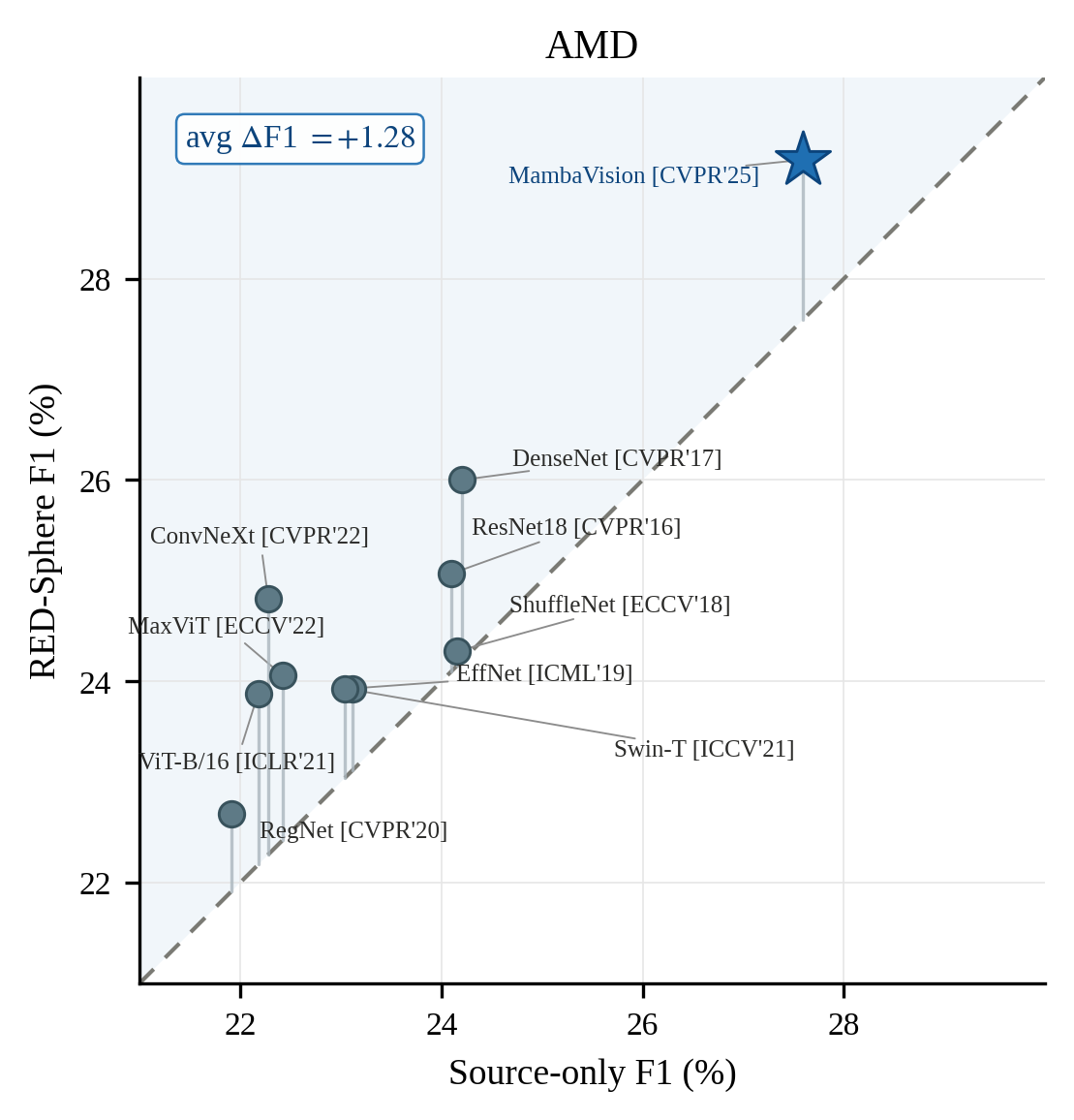}
  \end{minipage}\hspace{0.04\linewidth}%
  \begin{minipage}{0.35\linewidth}\centering
    \includegraphics[width=\linewidth]{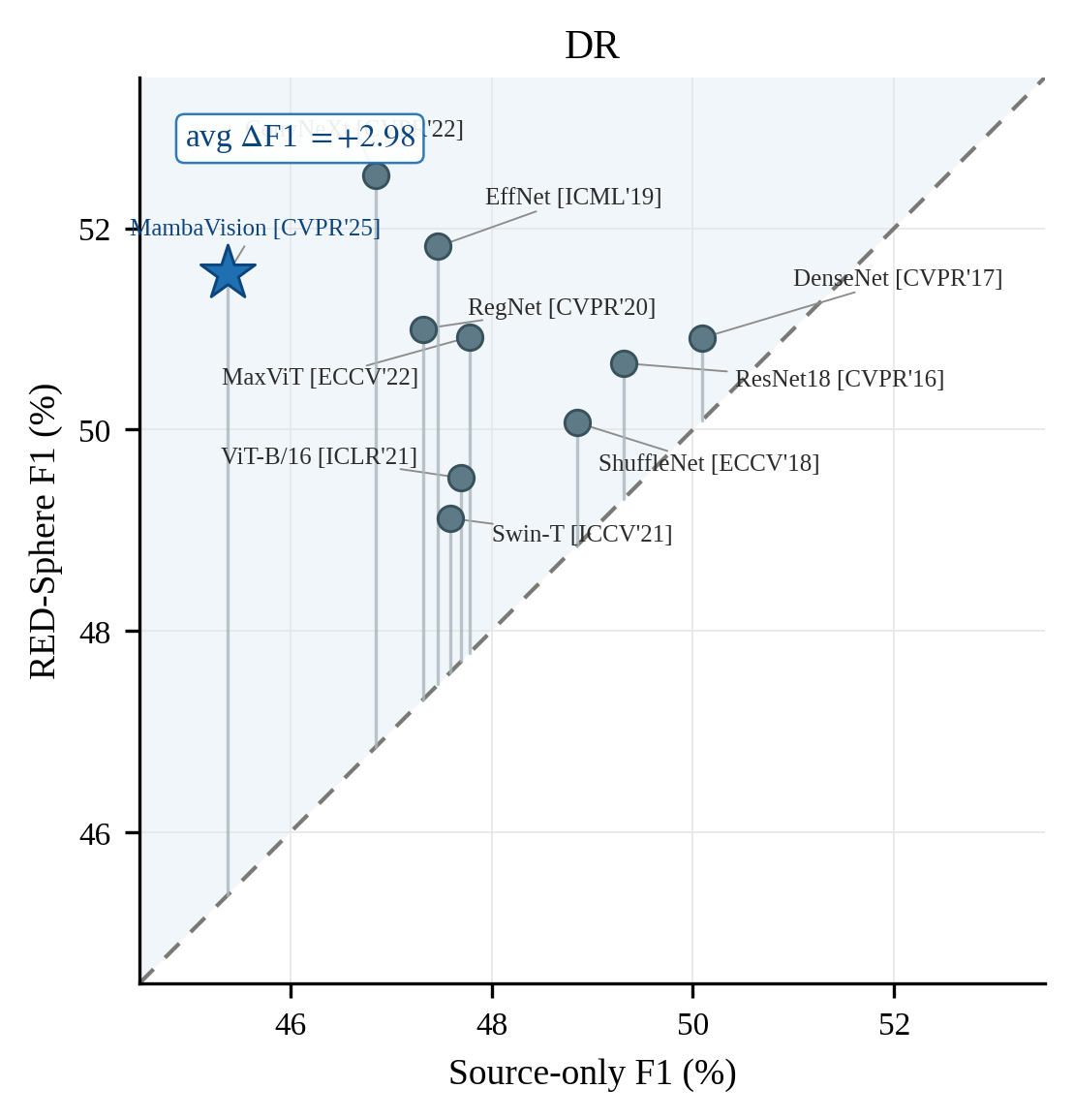}
  \end{minipage}
  \vspace{0.25cm}
  \caption{Teaser comparison between matched source only baselines and RED-Sphere on held out external cohorts. Points above the diagonal indicate improved external macro F1 after adding RED-Sphere to the same backbone. Stars mark MambaVision-T.}
  \label{fig:teaser}
  \vspace{0.1em}
\end{figure}

Figure~\ref{fig:teaser} provides a controlled summary of the empirical effect of RED-Sphere. The comparison keeps the backbone architecture, source cohort, external cohorts, and source-only model selection rule identical between the baseline and the proposed module, so that the observed differences can be attributed to the robustness mechanism itself. Under the White-only Harvard-FairVision 2D SLO protocol, Asian and Black cohorts are held out exclusively for external evaluation. Across ten representative visual backbones and two disease classification tasks, RED-Sphere consistently improves held-out macro-F1 in all 20 task-backbone settings, with average gains of about 1.3 points for AMD and 2.98 points for DR. The improvements are accompanied by higher average AUC and PR-AUC, suggesting stronger external ranking ability rather than a threshold-specific effect.

The main contributions are as follows:
\begin{itemize}
    \vspace{-0.1cm}
    \item We formulate a source-only cross-population generalization setting for medical image classification, where external populations are unavailable for training, validation, scheduling, hyperparameter selection, and model selection. The setting better reflects deployment where target population data are often inaccessible before use.
    \vspace{-0.1cm}
    \item We introduce a semantic-nuisance regularization strategy that constrains shortcut sensitive responses during training and, with hyperspherical prototype classification, promotes angular disease semantics over population correlated feature magnitude.
    \vspace{-0.1cm}
    \item We introduce a counterfactual-inspired nuisance regularization scheme that constructs masked nuisance views in feature space and constrains their predictions against the semantic stream, encouraging the classifier to remain stable when shortcut-sensitive responses are emphasized.
    \vspace{-0.1cm}
    \item We validate RED-Sphere on both AMD and DR datasets with multiple representative backbones, showing consistent effectiveness across datasets and architectures.
    \vspace{-0.5cm}
\end{itemize}

\section{Related Work}
\label{sec:related}

\vspace{-0.25cm}
\subsection{Source-Only Cross-Population Robustness in Ophthalmic Imaging}
\label{sec:related:sourceonly}

Medical imaging fairness has moved beyond post-hoc disparity reporting toward reliability under clinically meaningful demographic variation. Equality of opportunity and counterfactual fairness define influential group-aware criteria~\cite{hardt2016equality,kusner2017counterfactual}. Gichoya et al.~\cite{gichoya2022ai}, Glocker et al.~\cite{glocker2023algorithmic}, and Yang et al.~\cite{yang2024limits} further show that medical images can encode protected attributes even when such labels are hidden from the classifier. Ophthalmic benchmarks such as Harvard-FairVision~\cite{luo2024fairvision}, FairCLIP~\cite{luo2024fairclip}, FairDomain~\cite{tian2024fairdomain}, and FairMedFM~\cite{jin2024fairmedfm} show that fundus, SLO, OCT, and medical foundation models can exhibit demographic performance gaps. These observations make cross-population robustness central to screening, where source validation can underestimate failure on unseen patient populations.

Most mitigation strategies assume information unavailable in strict source-only deployment. FairTune adapts medical models with fairness-aware fine tuning~\cite{dutt2024fairtune}, FairQuantize studies quantization as a fairness intervention~\cite{fairquantize2024}, and slicing-based analyses expose hidden subgroups that can drive medical performance gaps~\cite{olesen2025slicing}. Domain generalization studies emphasize rigorous unseen-domain evaluation~\cite{gulrajani2021insearch}, while representative single-source methods broaden source variation through distribution uncertainty or feature-distribution matching~\cite{li2022dsu,zhang2022efdm}. Geirhos et al.~\cite{geirhos2019imagenet} show that ImageNet-trained CNNs can favor texture over shape, which motivates caution when fundus classifiers inherit low-level biases from pretrained backbones. Cross-population fundus shift is harder than ordinary appearance variation because pigmentation, illumination, vessel contrast, edge strength, and texture can correlate with population, while drusen, hemorrhages, exudates, vessel abnormalities, and lesion boundaries also depend on local structure. Broad perturbation therefore cannot decide which low-level responses are harmful shortcuts and which responses are diagnostic evidence.

RED-Sphere targets strict source-only deployment, where external populations are absent from all optimization and selection stages. Residual soft gating attenuates edge and feature-energy shortcuts, while a direct residual path preserves lesion structure. The goal is a deployable fundus classifier that weakens source-correlated appearance reliance without demographic supervision, rather than a stronger benchmark baseline.

\vspace{-0.25cm}
\subsection{Feature-Space Nuisance Learning and Hyperspherical Prediction}
\label{sec:related:nuisance_sphere}

Counterfactual reasoning provides a principled way to ask whether predictions remain stable when nuisance or sensitive factors change~\cite{kusner2017counterfactual}. Image-level counterfactuals and diffusion models can modify disease evidence, acquisition factors, or shortcut cues~\cite{vlontzos2023estimating,Weng2024diffusion,kumar2025prism}, but clinical use requires caution because generated anatomy or pathology can be difficult to audit. Cohen et al.~\cite{cohen2018hallucinate} show that distribution-matching medical image translation can hallucinate image features and fail to preserve task labels. A parallel line studies decision geometry: CosFace~\cite{wang2018cosface}, ArcFace~\cite{deng2019arcface}, and SphereFaceRevived~\cite{liu2022sphereface} normalize embeddings and classifiers so predictions depend on angular similarity rather than unbounded feature magnitude. Hyperspherical prototypes and neural-collapse analyses further motivate class-centered angular structure for discriminative representations~\cite{mettes2019hyperspherical,papyan2020neural}.

Feature-space alternatives avoid image synthesis. Liu et al.~\cite{liu2025imdr} learn disentangled ophthalmic representations for incomplete modalities, while Xia et al.~\cite{xia2024drdisentangle} separate diabetic-retinopathy semantics from domain noise. Wang et al.~\cite{wang2024navigate} connect shortcut debiasing to neural-collapse geometry, Zhang et al.~\cite{zhang2024dim} discover multiple biased subgroups, and Li et al.~\cite{li2025nsf} exploit sample clustering to mitigate spurious correlation without explicit bias labels. Prototype methods have also supported few-shot medical segmentation~\cite{zhang2024pcm} and biomedical vision-language prompting on a hyperspherical manifold~\cite{shao2026vmfcoop}. These approaches improve invariance or geometry, but most rely on observable variation, generated counterfactuals, subgroup discovery, or objectives designed for discriminability rather than strict source-only cross-population fundus deployment. In such deployment, feature magnitude may still encode illumination, contrast, image quality, or population-correlated appearance.

RED-Sphere couples nuisance regularization with angular prediction in a source-only module. An edge-dominant masked feature view is used only during training through consistency and separation losses, which reduce collapse between disease semantics and appearance-sensitive directions. At deployment, normalized disease prototypes classify the residual-gated stream by angular evidence rather than raw activation magnitude or target-population calibration.

\vspace{-0.5cm}
\section{Methodology}
\label{sec:method}

\vspace{-0.25cm}
\subsection{Problem Definition}
\label{sec:method:problem}

We study source-only cross-population fundus disease classification. Each sample contains a fundus image $x\in\mathbb{R}^{3\times H\times W}$, a disease label $y\in\{1,\ldots,K\}$, and a population attribute $a$. During training, however, only one source population is available for optimization and model selection. The population attribute is used solely to define the source and external evaluation partitions, instead of providing to the network as an input, an auxiliary target, a reweighting variable, or a validation criterion. This setting is stricter than standard demographic fairness learning: examples such as equality-of-opportunity or equalized-odds optimization~\cite{hardt2016equality}, counterfactual fairness~\cite{kusner2017counterfactual}, group-DRO~\cite{sagawa2020distributionally}, and demographic adversarial debiasing~\cite{zhang2018mitigating} require sensitive labels and multiple observed groups during training or validation so that group-conditional errors, risks, adversarial group predictors, or counterfactual criteria can be computed. Moreover, this setting differs from target-domain adaptation, where unlabeled or labeled target images may be observed before deployment. In our work, the target populations are unseen until external testing. The objective is therefore not to equalize observed demographic groups during training, but to learn disease-semantic features that preserve clinically meaningful retinal evidence while reducing dependence on source-correlated appearance cues that may shift across populations.

\vspace{-0.25cm}
\subsection{Framework Overview}
\label{sec:method:overview}

\begin{figure}[t]
    \centering
    \includegraphics[width=\linewidth]{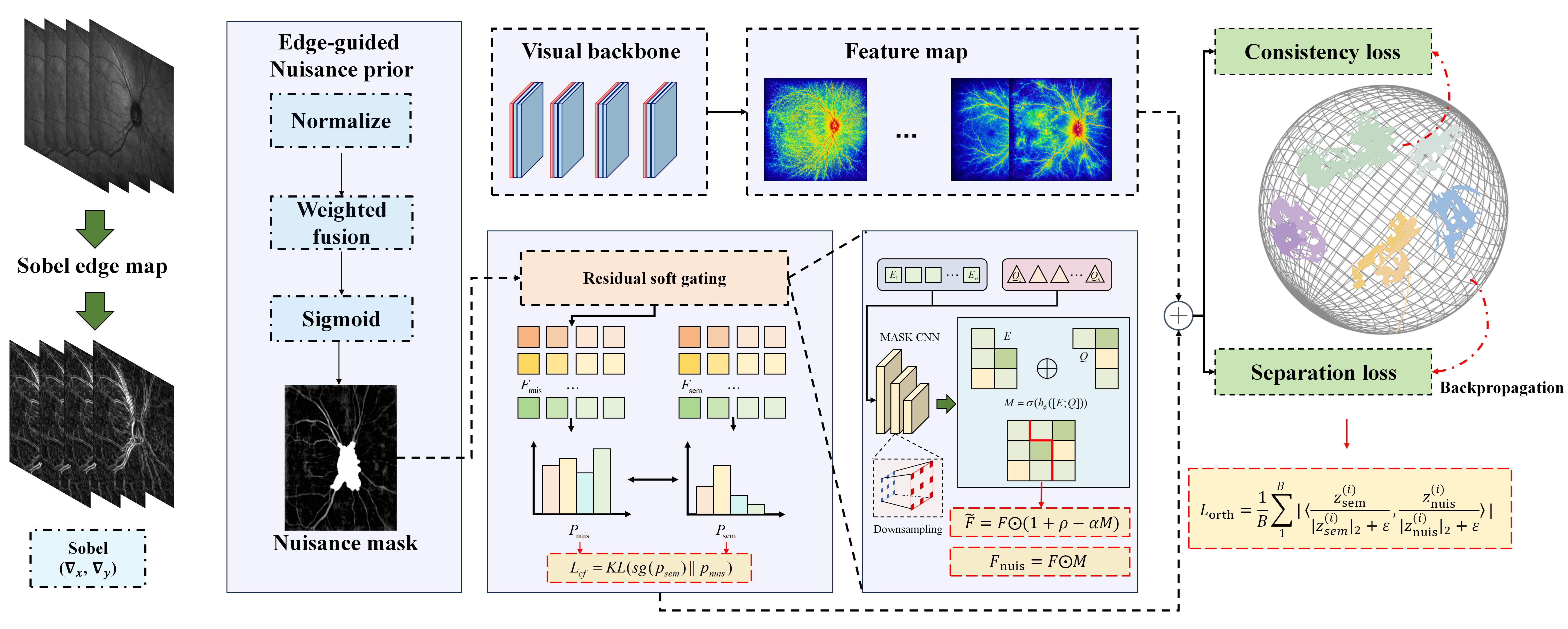}
    \vspace{0.1cm}
    \caption{Overview of the proposed RED-Sphere framework.}
    \label{fig:method_overview}
    \vspace{-0.1em}
\end{figure}

Figure~\ref{fig:method_overview} is the overview of the proposed RED-Sphere. A backbone encoder $B_{\theta}$ extracts a final feature map, and a feature transformer $\mathcal{T}_{\phi}$ decomposes it into a semantic stream, a nuisance stream, and a spatial mask:
\[
    F = B_{\theta}(x)\in\mathbb{R}^{C\times H\times W},
    \qquad
    (F_{\mathrm{sem}},F_{\mathrm{nuis}},M)=\mathcal{T}_{\phi}(x,F),
\]
where $C$ is the number of feature channels, $H\times W$ is the backbone feature resolution, $M\in(0,1)^{1\times H\times W}$ is a soft shortcut-sensitive mask, $F_{\mathrm{sem}}\in\mathbb{R}^{C\times H\times W}$ is the residual-gated semantic feature map used for disease prediction, and $F_{\mathrm{nuis}}\in\mathbb{R}^{C\times H\times W}$ is a masked nuisance view used only for training-time regularization. Whenever $M$ is multiplied with $F$, it is broadcast along the channel dimension.

The framework contains three coupled components. First, an edge-feature-energy estimator builds a lightweight nuisance cue from Sobel image-gradient responses~\cite{sobel1968isotropic} and local backbone feature energy. Second, residual soft gating attenuates mask-dominant responses while retaining a direct feature residual, avoiding hard deletion of lesion-relevant edge structure. Third, a spherical prototype classifier predicts disease labels from normalized semantic embeddings, making the decision depend on angular class evidence rather than raw feature magnitude. The design is backbone-agnostic: changing the visual backbone changes only $C$, and changing the task changes only the number of disease prototypes $K$.

\vspace{-0.25cm}
\subsection{Edge-Texture Guided Residual Debiasing}
\label{sec:method:rgfd}

Ophthalmic fairness and fundus domain-generalization studies show that retinal images can encode population information through pigmentation, image luminance, color statistics, and vessel-map information~\cite{luo2024fairvision,coyner2023racialbias,rajesh2025retinalpigment,jang2025fundusdg}. 
Motivated by these findings, we treat low-level factors such as retinal pigmentation, vessel-related contrast, illumination, texture, and edge strength as potential shortcut-sensitive cues.

These factors may correlate with the source population or acquisition process. They cannot, however, be blindly removed: drusen, hemorrhages, exudates, vessel abnormalities, and lesion boundaries are also locally structured and often edge-rich. RED-Sphere therefore uses edge and feature-energy responses to estimate where shortcut reliance is likely, and then applies residual gating rather than hard masking.

We first compute a Sobel-derived edge map from the input image. Let $g(x)\in\mathbb{R}^{1\times H_0\times W_0}$ denote the grayscale projection of the RGB input tensor, and let $S_x$ and $S_y$ be fixed horizontal and vertical Sobel kernels. For each image, the raw gradient-magnitude map is
\[
    A = \sqrt{(S_x * g(x))^2 + (S_y * g(x))^2 + \epsilon},
\]
where $*$ denotes convolution and $\epsilon>0$ is a numerical stabilizer. We use per-image linear min--max normalization. For any spatial map $R$, define
\[
    \operatorname{Norm}_{01}(R)_{uv}
    =\frac{R_{uv}-\min_{p,q}R_{pq}}
    {\max_{p,q}R_{pq}-\min_{p,q}R_{pq}+\epsilon}.
\]
The normalized edge map is resized to the backbone feature resolution by bilinear interpolation (Interp())\textbf{}:
\[
    E=\operatorname{Interp}_{H,W}\!\left(\operatorname{Norm}_{01}(A)\right)
    \in[0,1]^{1\times H\times W}.
\]

To complement image-level edges with feature-level activation concentration, we compute a local Gram-energy map
\[
    Q=\operatorname{Norm}_{01}\!\left(\frac{1}{C}\sum_{c=1}^{C}F_c\odot F_c\right)
    \in[0,1]^{1\times H\times W},
\]
where $F_c\in\mathbb{R}^{H\times W}$ is the $c$-th feature channel and $\odot$ denotes element-wise multiplication. The shortcut-sensitive mask is then predicted by a lightweight convolutional mask network $h_{\phi}:\mathbb{R}^{2\times H\times W}\rightarrow\mathbb{R}^{1\times H\times W}$:
\[
    M=\sigma\!\left(h_{\phi}([E;Q])\right),
\]
where $[E;Q]$ denotes channel-wise concatenation and $\sigma$ is the sigmoid function. Thus, $M$ is learned, but it is explicitly conditioned on two defined appearance-sensitive cues: image-gradient magnitude and local feature energy.

The mask controls a residual gate:
\[
    \widetilde{F}=F\odot(1-\alpha M)+\rho F
    =F\odot(1+\rho-\alpha M),
    \qquad
    \alpha=\operatorname{clip}(\alpha_{\mathrm{raw}},\alpha_{\min},\alpha_{\max}).
\]
Here $\alpha_{\mathrm{raw}}$ is a learnable scalar initialized by $\alpha_0$, $\alpha$ is its clipped value, and $\rho$ is the residual preservation coefficient. A high mask value therefore down-weights a response relative to low-mask regions, while the residual term prevents complete removal of retinal evidence. The gated feature is refined with a bottleneck residual refiner $r_{\phi}$ and channel context gate $c_{\phi}$:
\[
    U=\widetilde{F}+r_{\phi}(\widetilde{F}),
    \qquad
    F_{\mathrm{sem}}=U\odot c_{\phi}(U),
    \qquad
    F_{\mathrm{nuis}}=F\odot M.
\]
The context gate $c_{\phi}(U)\in(0,1)^{C\times1\times1}$ is broadcast spatially. The semantic stream $F_{\mathrm{sem}}$ is used for final prediction, whereas the nuisance view $F_{\mathrm{nuis}}$ contains mask-dominant responses and is used only during training.

\vspace{-0.25cm}
\subsection{Counterfactual-Inspired Nuisance Regularization}
\label{sec:method:counterfactual}

Because held-out populations are unavailable during training, RED-Sphere does not synthesize demographic-counterfactual fundus images or claim a causal intervention on population. We avoid relying on image-level translation for this purpose because distribution-matching medical image translation has been shown to hallucinate image features and may fail to preserve task labels, making generated images unsafe as direct clinical evidence~\cite{cohen2018hallucinate}. Instead, we use a feature-level counterfactual-inspired view: $F_{\mathrm{nuis}}$ emphasizes shortcut-sensitive responses selected by the learned mask, while $F_{\mathrm{sem}}$ retains residual-gated disease information. The nuisance stream is never used for deployment; it regularizes the decision geometry during training.

Semantic and nuisance streams are mapped into embeddings by separate projection heads:
\[
z_{\mathrm{sem}}=P_{\mathrm{sem}}(F_{\mathrm{sem}}),
    \qquad
    z_{\mathrm{nuis}}=P_{\mathrm{nuis}}(F_{\mathrm{nuis}}),
    \qquad
    z_{\mathrm{sem}},z_{\mathrm{nuis}}\in\mathbb{R}^{D}.
\]
The semantic embedding is used for the main prediction. The nuisance embedding is passed through the same classifier to produce an auxiliary prediction. Let $\ell_{\mathrm{sem}},\ell_{\mathrm{nuis}}\in\mathbb{R}^{K}$ be the corresponding logits and let $p_{\mathrm{sem}}=\operatorname{softmax}(\ell_{\mathrm{sem}})$ and $p_{\mathrm{nuis}}=\operatorname{softmax}(\ell_{\mathrm{nuis}})$. We impose a nuisance-view consistency loss
\[
    \mathcal{L}_{\mathrm{cf}}
    =\operatorname{KL}\!\left(\operatorname{sg}(p_{\mathrm{sem}})\,\|\,p_{\mathrm{nuis}}\right),
\]
where $\operatorname{sg}(\cdot)$ denotes stop-gradient. The semantic prediction acts as a fixed teacher for this term, forcing the nuisance view to remain prediction-compatible without making it the deployed classifier.

We also discourage collapse between the two embeddings by penalizing their normalized inner product. For a mini-batch of size $B$, the separation loss is
\[
    \mathcal{L}_{\mathrm{orth}}
    =\frac{1}{B}\sum_{i=1}^{B}
    \left|
    \left\langle
    \frac{z_{\mathrm{sem}}^{(i)}}{\|z_{\mathrm{sem}}^{(i)}\|_2+\epsilon},
    \frac{z_{\mathrm{nuis}}^{(i)}}{\|z_{\mathrm{nuis}}^{(i)}\|_2+\epsilon}
    \right\rangle
    \right|.
\]
The consistency term operates on class probabilities, while the separation term operates on representation directions. Together, they expose the classifier to nuisance-dominant perturbations without allowing the semantic and nuisance projections to collapse into the same feature direction.

\vspace{-0.3cm}
\subsection{Spherical Prototype Classification}
\label{sec:method:spherical}

Even after residual debiasing, population-correlated appearance can affect the magnitude of deep embeddings through illumination, contrast, image quality, or pigmentation-related activation strength. We therefore use a spherical prototype classifier, where prediction depends on angular similarity between normalized semantic embeddings and learned disease prototypes.

Let $w_c\in\mathbb{R}^{D}$ denote the prototype of class $c$. We normalize both the semantic embedding and prototypes:
\[
    \hat{z}=\frac{z_{\mathrm{sem}}}{\|z_{\mathrm{sem}}\|_2+\epsilon},
    \qquad
    \hat{w}_c=\frac{w_c}{\|w_c\|_2+\epsilon}.
\]
The cosine score for class $c$ is $a_c=\hat{z}^{\top}\hat{w}_c$. During training, we apply an additive class-aware cosine margin to the ground-truth class:
\[
    \ell_c=s\left(a_c-\mathbf{1}[c=y]m_y\right),
\]
where $s>0$ is a temperature, $\mathbf{1}[\cdot]$ is the indicator function, and $m_y$ is the margin of the ground-truth class. The margin is computed from the source training count $n_c$:
\[
    m_c=m_{\min}+(m_{\max}-m_{\min})\left(1-\sqrt{\frac{n_c}{\max_j n_j}}\right).
\]
Classes with fewer source examples receive larger margins, encouraging better angular separation for under-represented disease categories. At inference, no ground-truth label is available, so the additive margin is disabled and prediction uses cosine similarity to the normalized prototypes.

\vspace{-0.25cm}
\subsection{Training Objective and Inference}
\label{sec:method:training}

Training uses only source-population images. The semantic stream is optimized for disease classification with a class-balanced focal loss $\mathcal{L}_{\mathrm{cls}}$, while the nuisance stream contributes only auxiliary regularization. To avoid degenerate mask activation, we add a weak mean-mask penalty
\[
    \mathcal{L}_{\mathrm{mask}}
    =\frac{1}{BHW}\sum_{i=1}^{B}\sum_{u=1}^{H}\sum_{v=1}^{W}M_{iuv}.
\]
The complete objective is
\[
    \mathcal{L}
    =\mathcal{L}_{\mathrm{cls}}
    +\lambda_{\mathrm{cf}}\mathcal{L}_{\mathrm{cf}}
    +\lambda_{\mathrm{orth}}\mathcal{L}_{\mathrm{orth}}
    +\lambda_{\mathrm{mask}}\mathcal{L}_{\mathrm{mask}},
\]
where $\lambda_{\mathrm{cf}}$, $\lambda_{\mathrm{orth}}$, and $\lambda_{\mathrm{mask}}$ control nuisance-view consistency, embedding-direction separation, and mask sparsity, respectively. At inference, $P_{\mathrm{nuis}}$, $F_{\mathrm{nuis}}$, and all auxiliary losses are discarded. The model predicts from $F_{\mathrm{sem}}$ using the spherical prototype classifier and requires no target-population images, population attributes as model inputs, adaptation, or test-time calibration.

\vspace{-0.5cm}
\section{Experimental Results and Analysis}
\label{sec:exp}

\vspace{-0.25cm}
\subsection{Datasets and Metrics}
\label{sec:setup}

\paragraph{Datasets.}
We evaluate the AMD and DR cohorts of Harvard-FairVision~\cite{luo2024fairvision}, a public ophthalmic fairness benchmark with disease labels and demographic annotations. Harvard-FairVision provides both 2D SLO fundus images and 3D OCT B-scans. The experiments use only the 2D SLO fundus modality to isolate cross-population generalization in fundus-image classifiers. AMD is formulated as a four-class grading task with normal, early AMD, intermediate AMD, and late AMD labels. DR is formulated as a binary screening task that separates non-vision-threatening DR from vision-threatening DR. 
Both tasks are strongly class-imbalanced, which motivates the macro-averaged metrics described below.

\vspace{-0.25cm}
\paragraph{Protocol.}
We adopt a White-only source protocol as a deployment stress test for unseen populations. Model optimization uses only White samples from the official training split, and model selection uses only the White validation split. The White test split provides in-distribution (ID) evaluation. Asian and Black samples are excluded from training, validation, learning-rate scheduling, early stopping, hyperparameter selection, and model selection. Asian and Black samples are used exclusively as held-out cross-race (CR) test cohorts. Table~\ref{tab:fairvision_protocol} reports cohort sizes that define the evaluation partitions, while class imbalance motivates the metric design.

\vspace{-0.25cm}
\paragraph{Metrics.}

Macro-averaged F1 is the primary endpoint.
Accuracy is not used as the primary comparison metric under the proposed protocol. Non-vision-threatening DR accounts for 91.4\% of the White training cohort and 92.3\% of the Asian CR cohort, while Normal accounts for 92.9\% of the Black AMD cohort. 
We report macro-F1 as the main performance measure because macro-F1 gives equal weight to each disease class~\cite{he2009learning}. We also report AUC and PR-AUC as threshold-free ranking metrics. PR-AUC is especially informative when positive or severe disease cases are rare~\cite{saito2015precision}. 
For cross-population robustness, White F1 measures ID performance, and Asian and Black F1 measures held-out CR performance. External Avg F1 is the mean of Asian and Black F1. Worst External F1 is the lower value between Asian and Black F1. Cross-Race Gap is the absolute difference between Asian and Black F1, where a lower value indicates more balanced behavior across unseen racial groups. All splits are disjoint at the subject-ID level.

\begin{table*}[t!]
\centering
\small
\setlength{\tabcolsep}{7pt}
\renewcommand{\arraystretch}{1.10}
\vspace{0.5em}
\adjustbox{max width=\textwidth}{%
\begin{tabular}{@{}lrrrrrr@{}}
\toprule
\textbf{Task}
& \multicolumn{3}{c}{\textbf{White source cohort}}
& \multicolumn{3}{c}{\textbf{Held-out CR cohorts}} \\
\cmidrule(lr){2-4} \cmidrule(lr){5-7}
& \textbf{Train} & \textbf{Val.} & \textbf{ID Test}
& \textbf{Asian CR} & \textbf{Black CR} & \textbf{CR Total} \\
\midrule
AMD & 4{,}873 & 839 & 2{,}411 & 706 & 1{,}171 & 1{,}877 \\
DR  & 4{,}702 & 762 & 2{,}315 & 764 & 1{,}457 & 2{,}221 \\
\bottomrule
\end{tabular}}
\caption{Harvard-FairVision 2D SLO fundus cohort sizes under the White-only cross-race protocol, after modality filtering and protocol splitting. Asian and Black cohorts are reserved exclusively for held-out cross-race (CR) testing.}
\label{tab:fairvision_protocol}
\end{table*}

\vspace{-0.25cm}
\subsection{Implementation Details}

All images are resized to $224 \times 224$ and normalized with ImageNet statistics. Training uses random horizontal flipping, random rotation within $\pm 10^\circ$, mild color jittering, AdamW, batch size 32, an initial learning rate of $1\times10^{-6}$, and weight decay $1\times10^{-4}$. Validation and test transforms use only resizing and normalization.

Learning rate scheduling, checkpoint selection, and method specific hyperparameter calibration are controlled exclusively by White validation macro-F1. 
The same preprocessing, augmentation, and source only selection protocol are used for all baselines and backbones. RED-Sphere uses a 256 dimensional semantic embedding, hyperspherical temperature 16, class balanced focal loss, and fixed nuisance regularization coefficients across tasks and backbones. Each configuration is repeated over three seeds, and full hyperparameters are provided in the \textbf{supplementary material}.

\vspace{-0.25cm}
\subsection{Main Comparison and Backbone Generalization}

We evaluate RED-Sphere across ten image classification backbones to test whether the proposed robustness mechanism transfers across architectural families rather than matching a single inductive bias. The sweep covers classical CNNs (ResNet18~\cite{he2016resnet}, DenseNet121~\cite{huang2017densenet}), efficient CNNs (EfficientNet-B0~\cite{tan2019efficientnet}, RegNet-Y-400MF~\cite{radosavovic2020regnet}, ShuffleNet-V2~\cite{ma2018shufflenetv2}), Transformer architectures (ViT-B/16~\cite{dosovitskiy2021vit}, Swin-T~\cite{liu2021swin}), a modern ConvNet (ConvNeXt-Tiny~\cite{liu2022convnext}), a hybrid attention backbone (MaxViT-T~\cite{tu2022maxvit}), and a hybrid Mamba-Transformer backbone (MambaVision-T~\cite{hatamizadeh2025mambavision}). For each backbone, \emph{Base} denotes the matched source-only classifier, and \emph{Ours} denotes the same classifier equipped with RED-Sphere under the fixed source-only configuration.

\begin{table*}[tbp]
\centering
\renewcommand{\arraystretch}{1.0}
\setlength{\tabcolsep}{3.5pt}
\footnotesize
\vspace{0.2cm}   

\begin{tabular*}{\textwidth}{@{\extracolsep{\fill}}l ccc cc cc@{}}
\toprule
\multicolumn{8}{c}{\textbf{AMD dataset}} \\
\midrule
\multirow{2}{*}{\textbf{Model}}
& \multicolumn{3}{c}{\textbf{F1-Score}}
& \multicolumn{2}{c}{\textbf{AUC}}
& \multicolumn{2}{c}{\textbf{PR-AUC}} \\
\cmidrule(lr){2-4} \cmidrule(lr){5-6} \cmidrule(lr){7-8}
& Base & Ours & $\Delta$ & Base & Ours & Base & Ours \\
\midrule
ConvNeXt-Tiny
& 22.28$\pm$0.64 & \textbf{24.82$\pm$2.14} & +2.54
& 49.67$\pm$0.72 & \textbf{52.27$\pm$1.17}
& 25.23$\pm$0.30 & \textbf{26.39$\pm$0.77} \\
DenseNet121
& 24.20$\pm$1.21 & \textbf{26.00$\pm$0.76} & +1.80
& 50.09$\pm$2.68 & \textbf{58.73$\pm$1.92}
& 25.30$\pm$0.45 & \textbf{29.07$\pm$1.13} \\
EfficientNet-B0
& 23.12$\pm$0.78 & \textbf{23.93$\pm$1.23} & +0.81
& 48.77$\pm$3.00 & \textbf{52.86$\pm$0.57}
& \cellcolor{redbox}24.98$\pm$0.37 & \textbf{25.91$\pm$0.33} \\
MaxViT-T
& 22.42$\pm$0.68 & \textbf{24.06$\pm$0.24} & +1.64
& 49.14$\pm$0.59 & \textbf{52.20$\pm$1.68}
& 25.41$\pm$0.42 & \textbf{26.06$\pm$0.35} \\
RegNet-Y-400MF
& \cellcolor{redbox}21.91$\pm$2.34 & \cellcolor{yellowbox}\textbf{22.68$\pm$0.28} & +0.77
& \cellcolor{redbox}47.15$\pm$0.74 & \textbf{51.49$\pm$0.22}
& 25.03$\pm$0.17 & \cellcolor{yellowbox}\textbf{25.45$\pm$0.04} \\
ResNet18
& 24.10$\pm$0.16 & \textbf{25.07$\pm$0.23} & +0.97
& 49.98$\pm$0.37 & \textbf{58.62$\pm$0.86}
& 25.78$\pm$0.30 & \textbf{28.67$\pm$0.79} \\
ShuffleNet-V2-X1.0
& 24.16$\pm$0.43 & \textbf{24.30$\pm$0.86} & +0.14
& 49.47$\pm$0.25 & \textbf{51.86$\pm$1.73}
& 25.24$\pm$0.21 & \textbf{25.85$\pm$0.65} \\
Swin-T
& 23.04$\pm$0.04 & \textbf{23.93$\pm$0.82} & +0.89
& 49.36$\pm$0.50 & \cellcolor{yellowbox}\textbf{50.78$\pm$0.47}
& 25.12$\pm$0.19 & \textbf{25.88$\pm$0.16} \\
ViT-B/16
& 22.18$\pm$1.06 & \textbf{23.88$\pm$0.87} & +1.70
& 48.78$\pm$0.37 & \textbf{52.23$\pm$0.34}
& 25.37$\pm$0.40 & \textbf{25.62$\pm$0.78} \\
MambaVision-T
& \cellcolor{greenbox}27.59$\pm$0.44 & \cellcolor{bluebox}\textbf{29.18$\pm$0.34} & +1.59
& \cellcolor{greenbox}55.07$\pm$0.12 & \cellcolor{bluebox}\textbf{62.33$\pm$0.71}
& \cellcolor{greenbox}33.34$\pm$0.83 & \cellcolor{bluebox}\textbf{34.27$\pm$0.14} \\
\midrule
\textbf{Average}
& 23.50 & \textbf{24.79} & \textbf{+1.28}
& 49.75 & \textbf{54.34}
& 26.08 & \textbf{27.32} \\
\bottomrule
\end{tabular*}

\vspace{0.3em}

\begin{tabular*}{\textwidth}{@{\extracolsep{\fill}}l ccc cc cc@{}}
\toprule
\multicolumn{8}{c}{\textbf{DR dataset}} \\
\midrule
\multirow{2}{*}{\textbf{Model}}
& \multicolumn{3}{c}{\textbf{F1-Score}}
& \multicolumn{2}{c}{\textbf{AUC}}
& \multicolumn{2}{c}{\textbf{PR-AUC}} \\
\cmidrule(lr){2-4} \cmidrule(lr){5-6} \cmidrule(lr){7-8}
& Base & Ours & $\Delta$ & Base & Ours & Base & Ours \\
\midrule
ConvNeXt-Tiny
& 46.85$\pm$0.19 & \cellcolor{bluebox}\textbf{52.53$\pm$1.05} & +5.68
& \textbf{59.39$\pm$0.35} & 58.91$\pm$0.29
& \textbf{15.25$\pm$0.31} & 14.39$\pm$0.29 \\
DenseNet121
& \cellcolor{greenbox}50.09$\pm$1.32 & \textbf{50.91$\pm$0.35} & +0.82
& 51.56$\pm$0.18 & \textbf{58.24$\pm$1.21}
& 12.10$\pm$0.11 & \textbf{13.85$\pm$1.29} \\
EfficientNet-B0
& 47.47$\pm$0.45 & \textbf{51.82$\pm$0.06} & +4.35
& 55.06$\pm$2.70 & \textbf{56.28$\pm$0.98}
& 13.02$\pm$0.75 & \textbf{13.88$\pm$0.91} \\
MaxViT-T
& 47.78$\pm$1.21 & \textbf{50.92$\pm$0.71} & +3.14
& 58.01$\pm$0.80 & \textbf{60.23$\pm$1.30}
& 14.73$\pm$0.31 & \textbf{14.87$\pm$0.50} \\
RegNet-Y-400MF
& 47.32$\pm$0.25 & \textbf{51.00$\pm$0.72} & +3.68
& 53.05$\pm$1.23 & \textbf{57.24$\pm$0.62}
& 12.43$\pm$0.36 & \textbf{14.47$\pm$0.00} \\
ResNet18
& 49.31$\pm$0.73 & \textbf{50.66$\pm$3.95} & +1.35
& \cellcolor{redbox}49.14$\pm$1.61 & \textbf{60.14$\pm$0.16}
& \cellcolor{redbox}11.34$\pm$0.49 & \textbf{16.08$\pm$0.10} \\
ShuffleNet-V2-X1.0
& 48.85$\pm$1.24 & \textbf{50.07$\pm$2.50} & +1.22
& 51.46$\pm$0.65 & \cellcolor{yellowbox}\textbf{56.10$\pm$0.25}
& 11.77$\pm$0.23 & \cellcolor{yellowbox}\textbf{13.73$\pm$0.51} \\
Swin-T
& 47.59$\pm$0.62 & \cellcolor{yellowbox}\textbf{49.12$\pm$0.23} & +1.53
& 59.19$\pm$0.23 & \textbf{60.08$\pm$1.42}
& \textbf{14.93$\pm$0.05} & 14.89$\pm$0.77 \\
ViT-B/16
& 47.70$\pm$2.14 & \textbf{49.52$\pm$2.63} & +1.82
& 58.50$\pm$0.78 & \textbf{58.79$\pm$0.30}
& 14.33$\pm$0.51 & \textbf{14.64$\pm$0.01} \\
MambaVision-T
& \cellcolor{redbox}45.38$\pm$0.33 & \textbf{51.55$\pm$0.12} & +6.17
& \cellcolor{greenbox}60.00$\pm$0.67 & \cellcolor{bluebox}\textbf{68.73$\pm$0.63}
& \cellcolor{greenbox}15.81$\pm$0.87 & \cellcolor{bluebox}\textbf{19.15$\pm$0.20} \\
\midrule
\textbf{Average}
& 47.83 & \textbf{50.81} & \textbf{+2.98}
& 55.54 & \textbf{59.47}
& 13.57 & \textbf{15.00} \\
\bottomrule
\end{tabular*}
\caption{Cross-population generalization on AMD and DR. Macro-F1, AUC, and PR-AUC (\%) are averaged across the held-out Asian and Black cohorts. $\Delta$ denotes the macro-F1 gain of RED-Sphere over the matched source-only baseline. Bold marks the stronger value within each backbone comparison. For extrema within each column, \protect\colorbox{greenbox}{green} and \protect\colorbox{redbox}{red} mark the best and worst \emph{Base} values, whereas \protect\colorbox{bluebox}{blue} and \protect\colorbox{yellowbox}{yellow} mark the best and worst \emph{Ours} values.}
\label{tab:cross_pop_full}

\vspace{-0.1cm} 
\end{table*}

Table~\ref{tab:cross_pop_full} shows that RED-Sphere improves external disease recognition across backbone families rather than benefiting one particular architecture. Held-out macro-F1 increases in all 20 comparisons across the two tasks and ten backbones. Averaged over backbones, AMD external F1 improves from 23.50 to 24.79, a gain of about 1.3 F1 points, while DR external F1 improves from 47.83 to 50.81, a gain of 2.98 F1 points. The improvement also appears in ranking metrics. Average AUC increases by 4.59 points on AMD and 3.93 points on DR, while PR-AUC increases by 1.24 and 1.43 points, respectively. 
Because the Asian and Black cohorts are never used for any selection stage (\S\ref{sec:setup}), the gains reflect source-only external generalization rather than target-tuned selection.

The stronger DR gains are clinically meaningful because vision-threatening DR is rare, which makes source only decision boundaries fragile under population shift. RED-Sphere improves ConvNeXt-Tiny and MambaVision-T by 5.68 and 6.17 F1 points on DR, respectively. ResNet18 also raises external AUC from 49.14 to 60.14, moving from near chance behavior to clear discrimination. AMD shows smaller but more uniform gains, with every backbone improving in F1. The largest AMD AUC gains occur on DenseNet121 and ResNet18, both by 8.64 points, which suggests that residual edge debiasing benefits both dense feature reuse and residual feature propagation. The few metric level tradeoffs remain limited. DR ConvNeXt-Tiny gains 5.68 F1 points while AUC and PR-AUC decrease by 0.48 and 0.86 points, which indicates an operating point shift under class imbalance rather than a systematic loss of discrimination. 
Overall, Table~\ref{tab:cross_pop_full} supports RED-Sphere as a general robustness layer that reduces reliance on population sensitive appearance cues while preserving disease semantic separability.

\vspace{-0.25cm}
\subsection{Visual Analysis}

The visual diagnostics assess whether numerical gains correspond to cross-population semantic alignment, angular disease organization, and stable decision margins under external cohort shifts.

\begin{figure}[t]
  \centering
  \includegraphics[width=0.8\linewidth]{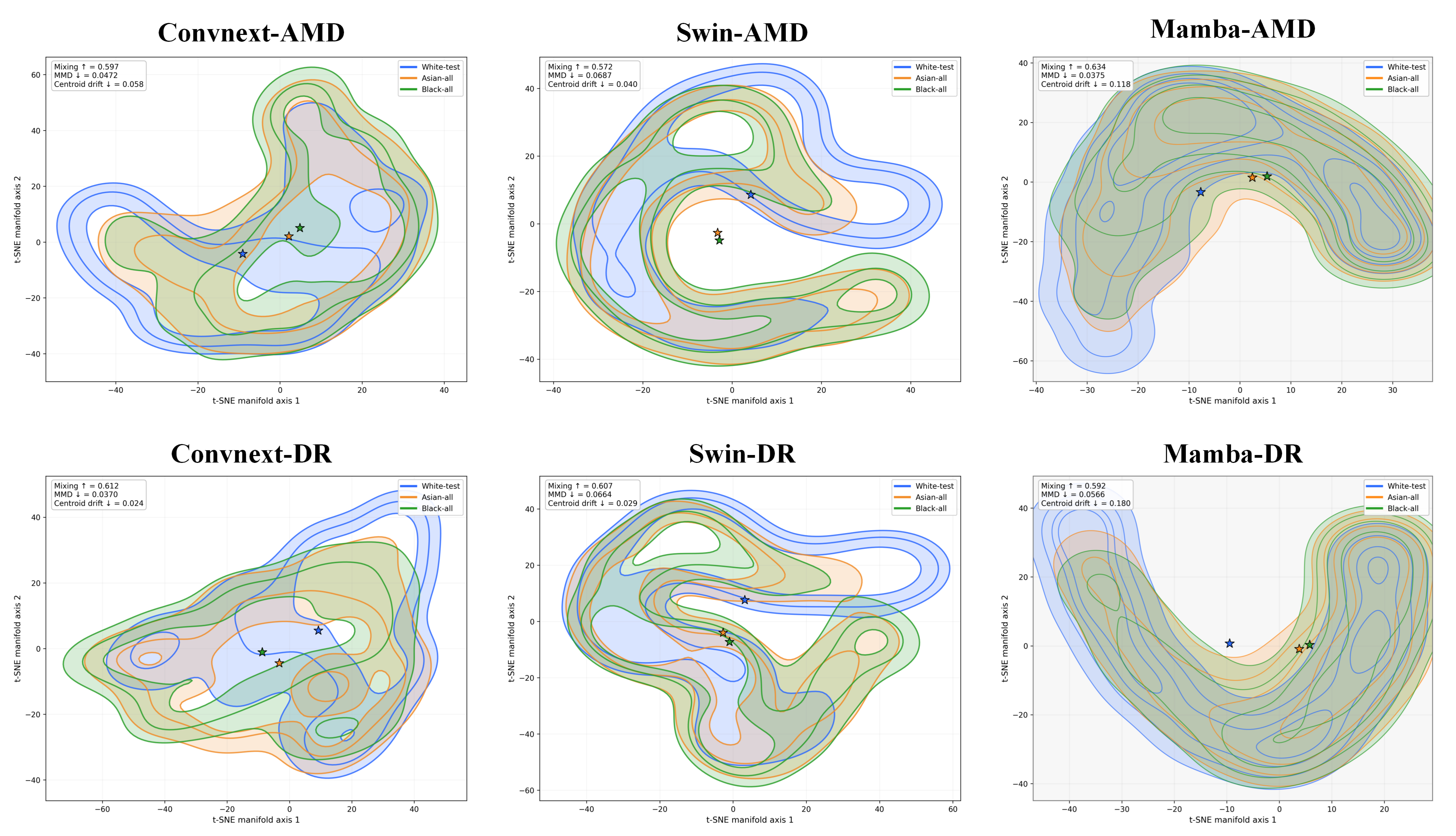}
  \vspace{0.2cm}
  \caption{Cross-cohort semantic manifold overlap under RED-Sphere on ConvNeXt-Tiny, Swin-T, and MambaVision-T. Kernel density contours are computed from semantic embeddings of the White test cohort and the held out Asian and Black cohorts. Mix, MMD, and centroid drift are reported in each panel. Higher Mix and lower MMD or centroid drift indicate stronger overlap between source and external cohorts. Centroid drift measures the distance between White and external cohort centroids in the projected semantic space.}
  \label{fig:density_overlap}
\end{figure}

\vspace{-0.25cm}
\paragraph{Cross cohort semantic overlap.}

Figure~\ref{fig:density_overlap} visualizes kernel density contours of semantic embeddings from the White test cohort and the held out Asian and Black cohorts. The density diagnostic asks whether unseen populations remain within the source semantic support after RED-Sphere training. Higher Mix scores and lower MMD values, which are printed above each panel, indicate stronger source to external overlap. 
Across ConvNeXt-Tiny, Swin-T, and MambaVision-T, the external contours generally share semantic support with the White contour rather than forming disjoint regions. The MambaVision panels show larger centroid drift, especially on DR, but the Asian and Black contours remain close to each other, which suggests that RED-Sphere reduces source specific appearance reliance while keeping both unseen cohorts behaviorally similar.

\begin{figure*}[t]
  \centering
  \includegraphics[width=\linewidth]{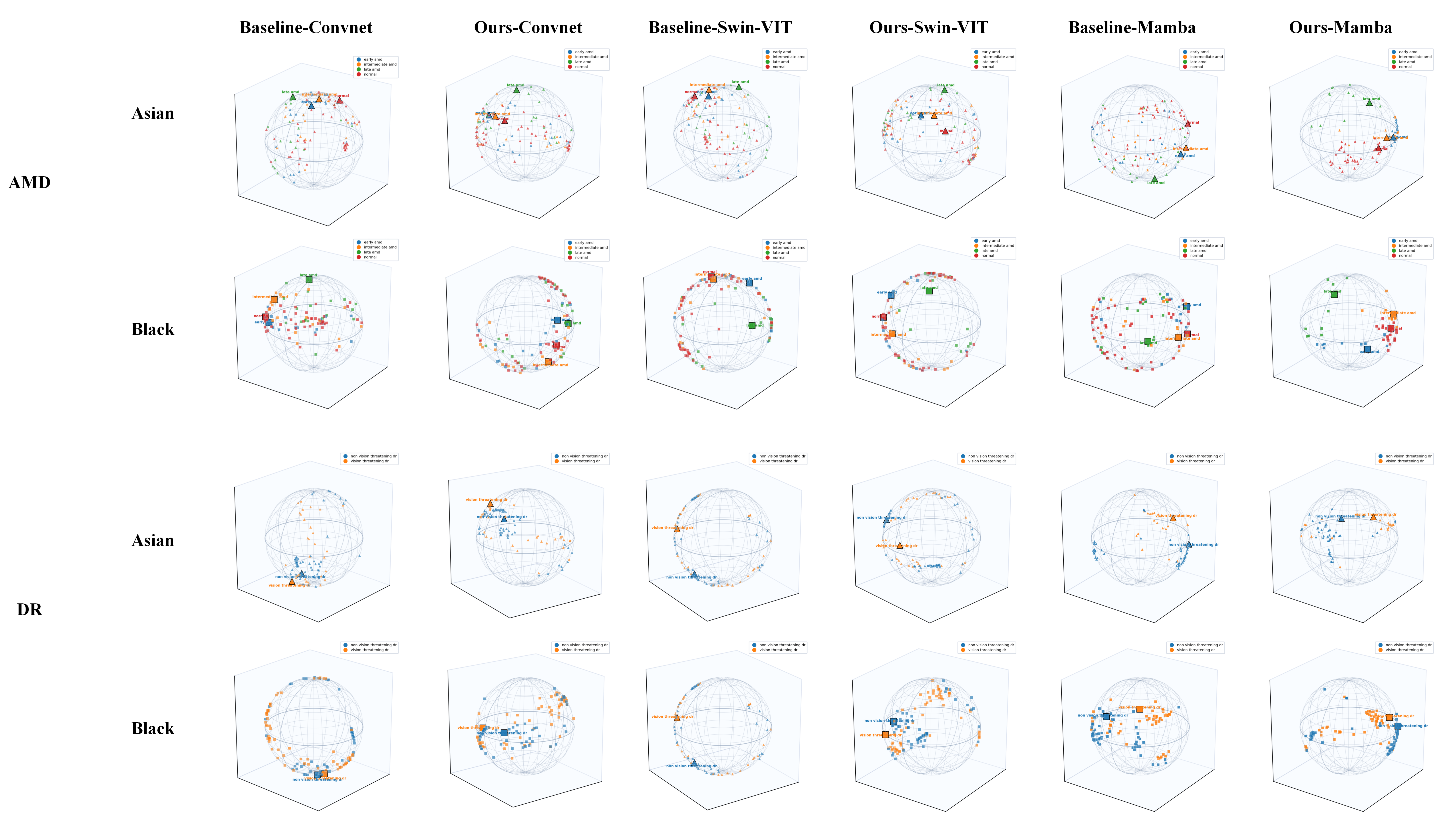}
  \vspace{0.1em}
  \caption{Spherical decision geometry on AMD and DR for ConvNeXt-Tiny, Swin-T, and MambaVision-T. Columns compare the matched source-only baseline and RED-Sphere on the unit hypersphere. Colors denote disease labels, and marker styles denote source and external cohorts. Coherent angular class regions indicate disease semantic separation under cross-population shift.}
  \label{fig:spherical_domain_clustering_3d}
  \vspace{-0.1cm}
\end{figure*}

\vspace{-0.25cm}
\paragraph{Spherical decision geometry.}
Figure~\ref{fig:spherical_domain_clustering_3d} projects normalized semantic embeddings onto the unit hypersphere. The spherical prototype classifier predicts by angular similarity to disease prototypes, so coherent class regions on the hypersphere provide direct evidence for disease semantic organization. RED-Sphere produces more compact and better separated angular regions than the matched source only baseline across AMD and DR. The baseline shows more diffuse class geometry, which suggests stronger dependence on feature magnitude and appearance driven activation strength. RED-Sphere preserves class structure more consistently across White, Asian, and Black cohorts, which supports the intended role of angular prediction in limiting population sensitive appearance effects.

\begin{figure*}[t]
  \centering
  \includegraphics[width=0.48\textwidth]{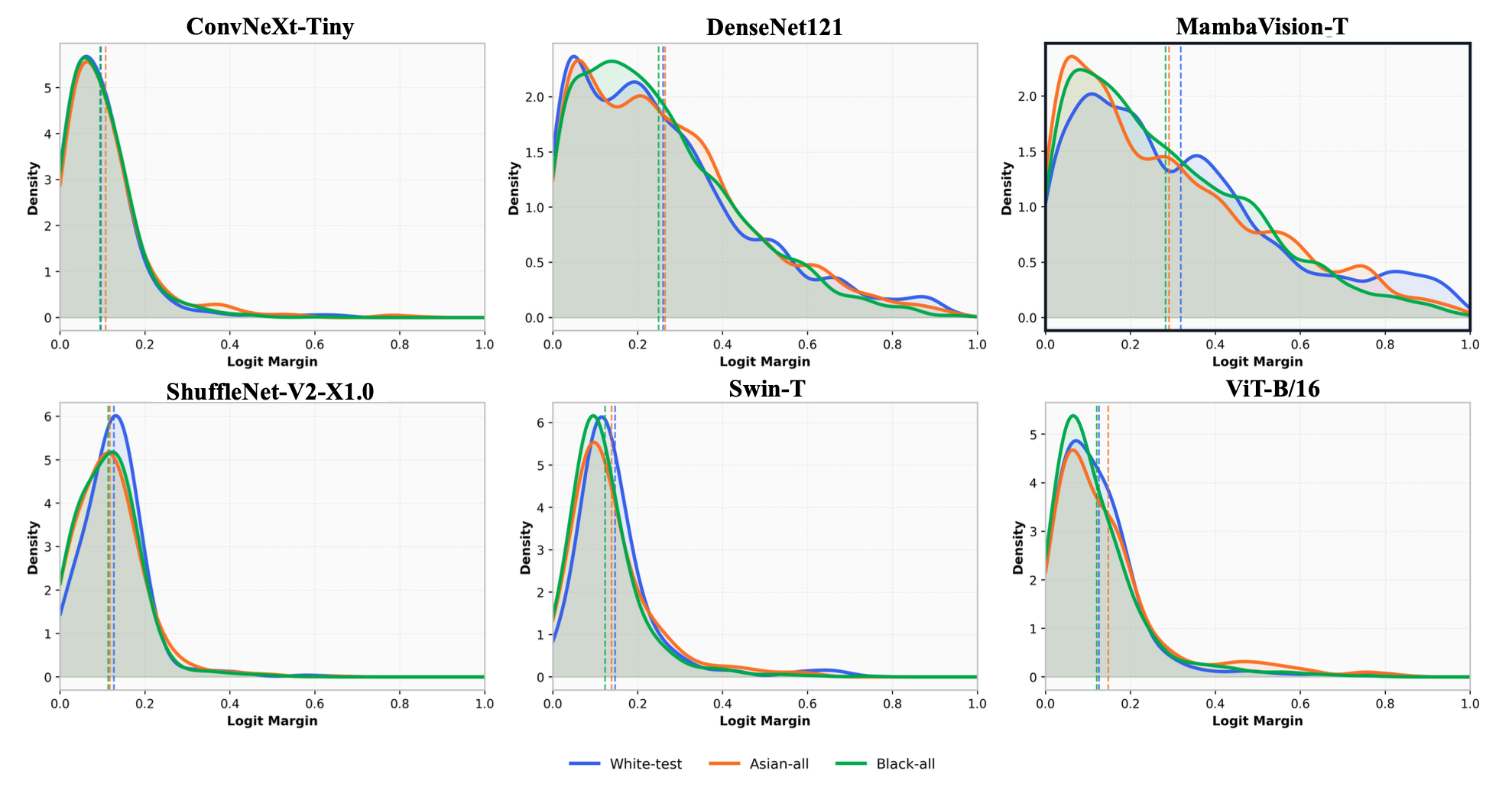}
  \hfill
  \includegraphics[width=0.48\textwidth]{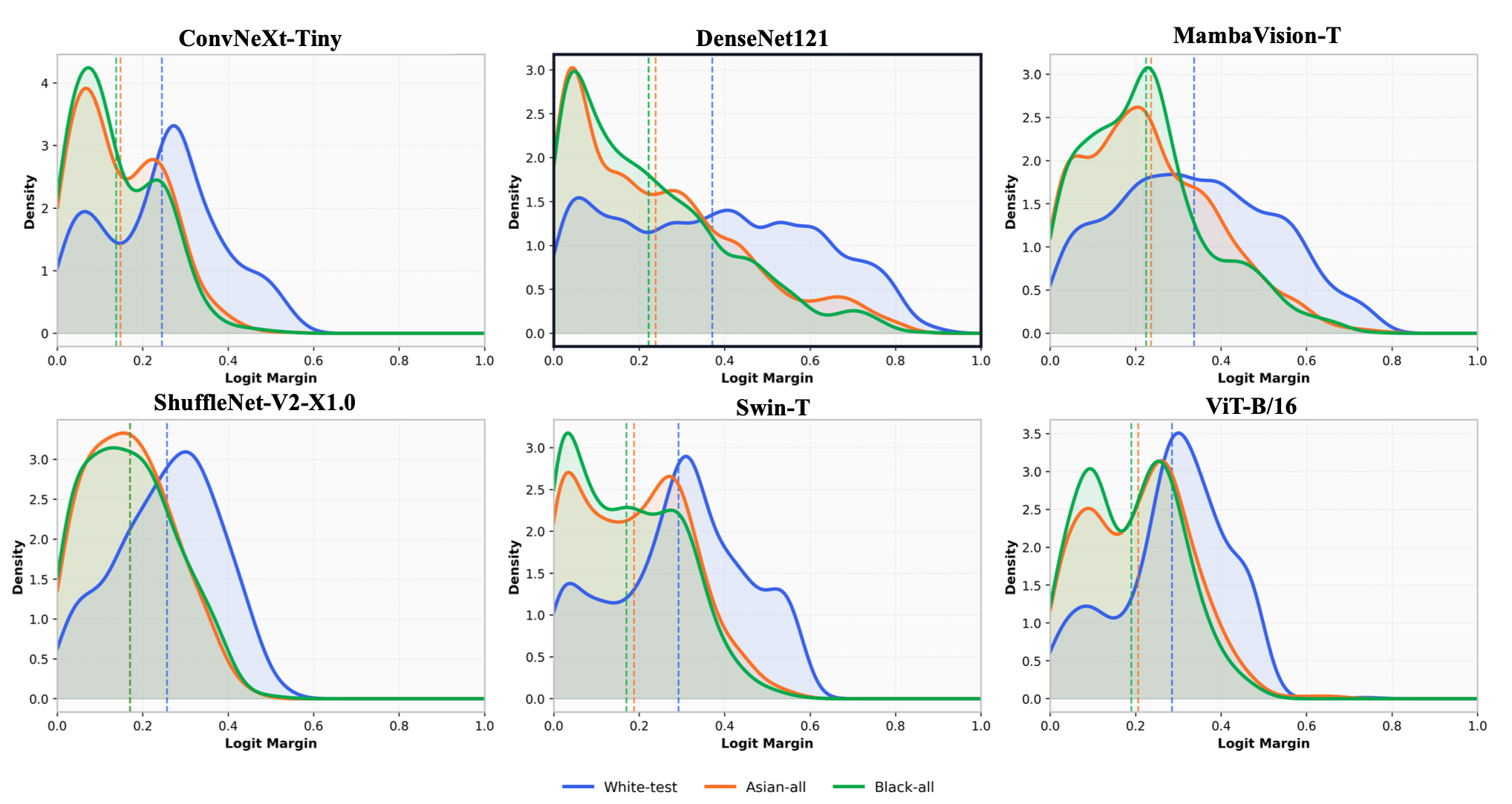}
    \vspace{0.3cm}
  \caption{Logit margin densities under RED-Sphere across White, Asian, and Black cohorts. Left: AMD. Right: DR. Dashed vertical lines denote cohort means. Closer curves indicate more stable decision boundaries across populations, while closer external cohort curves indicate more stable behavior under unseen population testing.}
  \label{fig:margin_density}
\end{figure*}

\vspace{-0.25cm}
\paragraph{Decision margin stability.}
Figure~\ref{fig:margin_density} examines decision boundary stability. For image $x$ with logits $\ell(x)$ and predicted class $\hat{y}$, the logit margin $d(x)=\ell_{\hat{y}}(x)-\max_{c\neq \hat{y}}\ell_c(x)$ measures the gap to the closest competing class and is used as a boundary diagnostic rather than calibrated clinical confidence. On AMD, White, Asian, and Black margin distributions remain closely aligned across representative backbones. On DR, White margins are larger for several backbones, reflecting the easier in distribution binary screening setting, while Asian and Black curves remain close. The remaining margin gap mainly reflects source versus external deployment difficulty rather than race specific divergence between unseen cohorts. Together, the density, sphere, and margin visualizations show semantic alignment, angular disease organization, and comparable external decision behavior.

\vspace{-0.25cm}
\subsection{Ablation Study}

Table~\ref{tab:ablation_amd_dr_retained_components} evaluates the three retained components of RED-Sphere. Removing the counterfactual inspired nuisance branch gives the largest mean signed change, $-1.18$ F1 points, with the largest losses on DR ConvNeXt and AMD ConvNeXt, showing that the branch turns residual gating into shortcut debiasing rather than generic spatial attention. Removing the class aware adaptive margin changes performance by $-0.96$ F1 points on average, supporting its role in expanding angular separation for rare disease categories. Removing the spherical prototype classifier degrades all six settings, with a mean signed change of $-0.89$ F1 points, showing that angular prediction stabilizes the decision space used by nuisance regularization and adaptive margins. The only local exception is AMD Swin without the nuisance branch, which changes by $+0.14$ F1 points and remains small relative to the reported standard deviations. The full model gives the best macro-F1 in five of six settings and remains effectively tied on AMD Swin.

\begin{table*}[htbp]
\centering
\renewcommand{\arraystretch}{1.18}
\vspace{0.3cm}
\setlength{\tabcolsep}{6pt}
\resizebox{\textwidth}{!}{
\begin{tabular}{lcccccc}
\hline
\textbf{Variant}
& \textbf{AMD-ConvNeXt}
& \textbf{AMD-Swin}
& \textbf{AMD-Mamba}
& \textbf{DR-ConvNeXt}
& \textbf{DR-Swin}
& \textbf{DR-Mamba} \\
\hline
Ours
& \textbf{24.82$\pm$2.14}
& \textbf{23.93$\pm$0.82}
& \textbf{29.18$\pm$0.34}
& \textbf{52.53$\pm$1.05}
& \textbf{49.12$\pm$0.23}
& \textbf{51.55$\pm$0.12} \\
\hline
w/o Adaptive Margin
& \makecell{23.47$\pm$1.91\\($-$1.35)}
& \makecell{23.86$\pm$0.74\\($-$0.07)}
& \makecell{27.90$\pm$1.29\\($-$1.28)}
& \makecell{50.94$\pm$0.88\\($-$1.59)}
& \makecell{48.33$\pm$0.31\\($-$0.79)}
& \makecell{50.88$\pm$0.17\\($-$0.67)} \\
w/o Counterfactual Branch
& \makecell{23.18$\pm$2.37\\($-$1.64)}
& \makecell{24.07$\pm$1.06\\(+0.14)}
& \makecell{27.94$\pm$0.35\\($-$1.24)}
& \makecell{50.21$\pm$1.34\\($-$2.32)}
& \makecell{47.68$\pm$1.46\\($-$1.44)}
& \makecell{50.95$\pm$0.21\\($-$0.60)} \\
w/o Spherical-Space Classifier
& \makecell{23.96$\pm$1.58\\($-$0.86)}
& \makecell{23.11$\pm$0.68\\($-$0.82)}
& \makecell{28.68$\pm$0.95\\($-$0.50)}
& \makecell{51.36$\pm$0.97\\($-$1.17)}
& \makecell{48.04$\pm$0.27\\($-$1.08)}
& \makecell{50.63$\pm$0.11\\($-$0.92)} \\
\hline
\end{tabular}
}
\caption{Component ablations on AMD and DR using three representative backbones. Macro-F1 (\%) is reported as mean$\pm$std. The value in parentheses reports the signed F1 change relative to the full RED-Sphere model.}
\label{tab:ablation_amd_dr_retained_components}
\vspace{-0.15cm}
\end{table*}

\vspace{-0.25cm}
\subsection{Sensitivity Analysis}

We further analyze the two hyperparameters that control residual preservation and initial shortcut suppression. Full results are in the \textbf{supplementary material}.
The residual preservation coefficient $\rho=0.50$ gives the best F1 in all six tasks and backbone settings. Smaller values can suppress lesion evidence together with population sensitive edge and texture responses, while larger values make the residual path dominate and weaken debiasing. The initial suppression coefficient $\alpha_0=0.20$ gives the best F1 in five of six settings. The only exception is DR Swin, where $\alpha_0=0.30$ improves F1 from 49.12 to 49.81. Stronger suppression degrades most settings, including AMD Swin and DR Swin at $\alpha_0=0.50$. We therefore use $\rho=0.50$ and $\alpha_0=0.20$ as fixed settings rather than tuning per backbone.

\vspace{-0.5cm}
\section{Conclusion}
\label{sec:conclusion}

\vspace{-0.15cm}
We presented RED-Sphere, a source-only framework for cross-population fundus disease domain generalization. It reduces reliance on source-correlated appearance cues through residual edge debiasing, counterfactual-inspired nuisance regularization, and spherical prototype classification. Under the White-only Harvard-FairVision SLO protocol, RED-Sphere improves held-out macro-F1 across all 20 task and backbone comparisons, with average gains of about 1.3 F1 points on AMD and 2.98 F1 points on DR. 
Visual analyses show shared semantic support, coherent angular disease regions, and stable external decision margins, while ablations confirm complementary roles for nuisance regularization, adaptive margins, and spherical prediction.

The broader contribution of RED-Sphere is a source-only robustness principle, estimating shortcut-sensitive cues, preserving evidence through residual modulation, regularizing nuisance-dominant representations, and predicting in normalized angular space, rather than a detector tied to one anatomy or modality. 
By defining task-specific nuisance priors, it can extend to other diseases, modalities, institutions, and acquisition protocols and to non-medical settings where appearance shortcuts entangle with semantic evidence.

\bibliography{egbib}
\end{document}